\documentclass[preprint]{vgtc}               % preprint
\ifpdf%                                % if we use pdflatex
  \pdfoutput=1\relax                   % create PDFs from pdfLaTeX
  \usepackage{graphicx}                % allow us to embed graphics files
  \DeclareGraphicsExtensions{.pdf,.png,.jpg,.jpeg} % for pdflatex we expect .pdf, .png, or .jpg files
\else%                                 % else we use pure latex
  \usepackage{graphicx}                % allow us to embed graphics files
  \DeclareGraphicsExtensions{.eps}     % for pure latex we expect eps files
\fi%

\graphicspath{{figures/}{pictures/}{images/}{./}} % where to search for the images

\usepackage{microtype}                 % use micro-typography (slightly more compact, better to read)
\PassOptionsToPackage{warn}{textcomp}  % to address font issues with \textrightarrow
\usepackage{textcomp}                  % use better special symbols
\usepackage{mathptmx}                  % use matching math font
\usepackage{times}                     % we use Times as the main font
\usepackage{cite}                      % needed to automatically sort the references
\usepackage{tabu}                      % only used for the table example
\usepackage{booktabs}                  % only used for the table example
\usepackage{multirow}
\usepackage{enumitem}

\onlineid{1070}

\vgtccategory{Empirical Study}

\vgtcinsertpkg

\title{LineCap: Line Charts for Data Visualization Captioning Models}

\author{Anita Mahinpei\thanks{e-mail: amahinpei@g.harvard.edu} %
\and Zona Kostic\thanks{e-mail: zonakostic@g.harvard.edu} %
\and Chris Tanner\thanks{e-mail: christanner@g.harvard.edu}}
\affiliation{\scriptsize Harvard University}

\abstract{Data visualization captions help readers understand the purpose of a visualization and are crucial for individuals with visual impairments. The prevalence of poor figure captions and the successful application of deep learning approaches to image captioning motivate the use of similar techniques for automated figure captioning. However, research in this field has been stunted by the lack of suitable datasets. We introduce \textit{LineCap}, a novel figure captioning dataset of 3,528 figures, and we provide insights from curating this dataset and using end-to-end deep learning models for automated figure captioning.} % end of abstract


\begin{document}


%%\firstsection{Introduction}

\maketitle

\section{Introduction} %for journal use above \firstsection{..} instead
Data visualizations are commonly used in scientific papers to convey complementary information and enhance readers' comprehension. Figure captions can help readers understand the purpose of a visualization, and they are often the only means for individuals with visual impairments to access figures. While guidelines for creating accessible visualizations exist, many authors have yet to adopt these practices when writing figure captions \cite{Lundgard:2022}. As such, automatic caption generation for data visualizations can significantly alleviate information inaccessibility for people with visual impairments. 

Early efforts for automated figure captioning have primarily focused on developing rule-based and non-deep learning techniques with modular pipelines \cite{corio1999, mittal1998, moraes2014, alzaidy2016, demir2010}. While these methods do not need large corpora of figures and captions, they are highly specialized and do not generalize well to other chart types and styles, or to charts that require more complex high-level insights. 

The advancement of deep learning has led to significant improvements in the field of automated image captioning \cite{li:2019}. Researching novel neural network architectures for image caption generation would not have been possible without the curation of publicly-shared datasets used for training and evaluation of these models. These datasets tend to contain large amounts of images scraped from the web with several captions crowd-sourced for each image. The Flickr30K dataset \cite{Flickr30}, the Microsoft COCO dataset \cite{MSCOCO}, and the VizWiz-Captions dataset \cite{vizwiz} are just a few examples of many image captioning datasets that have been created over the past decade. 

The recent successes of deep learning approaches in automated image captioning motivate the application of similar approaches to the task of automated figure captioning. However, research in this field has been stunted by the lack of suitable training and evaluation datasets. Therefore, we introduce \textit{LineCap}, a dataset containing line charts scraped from scientific papers each accompanied with crowd-sourced natural language descriptions. We share our design choices and challenges while curating LineCap, to help inform future creators of figure captioning datasets. We also establish baseline performances on LineCap and provide insights toward using deep learning models for automated figure captioning.

\section{Related Work}

Previous research has developed neural networks that generate natural language descriptions for charts when provided with the underlying data in tabular form \cite{obeid2020, chen2020, totto2020, truce2021, spreafico2020}. However, these data-to-text models and their accompanying datasets are not suitable for generating captions when the underlying figure data are not available, as is the case with most figures in scientific papers. Although Obeid et al. \cite{obeid2020} provide the original chart images in addition to data tables and captions, their data were crawled from Statista, \footnote{https://www.statista.com/} which only uses a limited set of chart styles and color schemes.

 Chen et al. \cite{figcap} and Qian et al. \cite{captionunits} use deep learning approaches to generate figure captions from chart images; however, these works trained models using FigureQA \cite{figureqa} and DVQA \cite{dvqa}, which are \textit{synthetic} figure question answering datasets. To generate reference captions for model training and evaluation, these works used a set of templates to create captions based on the question-answer pairs in the figure question answering datasets. The use of synthetic charts and template-based reference captions drastically limits the complexity of these datasets; the captions in these datasets only convey low-level information (e.g., chart type, axis titles, and global extrema) rather than high-level insights or trends. While describing low-level details is also important to visually impaired individuals, most low-level details can be extracted using figure parsing \cite{poco:2017, figureseer} or question answering models \cite{stlcqa, bimodalfusion} and incorporated into captions using natural language models. Furthermore, Lundgard et al. \cite{Lundgard:2022} suggests that automated data visualization captioning research should primarily focus on describing overall trends and statistics. They categorize figure caption information into four broad groups: Level 1 (e.g., chart type, labels, and axis ranges), Level 2 (e.g., descriptive statistics and extrema), Level 3 (e.g., complex trends and exceptions), and Level 4 (e.g., domain specific insights and explanations). Through a user study, they found that blind readers consistently rank level 2 and 3 information as most useful \cite{Lundgard:2022}. However, captions provided by Chen et al. \cite{figcap} and Qian et al. \cite{captionunits} commonly fall under level 1 and occasionally level 2.
 
 To address the limitations of synthetic data, the research community has introduced non-synthetic datasets including SciCap \cite{scicap}, ChaTa+ \cite{chata}, and Chart-to-Text \cite{chart2text}. SciCap is a dataset of figures and captions extracted from computer science papers published on arXiv between 2010 and 2020. ChaTa+ is a much smaller dataset of only 1,640 figures and captions extracted from scientific articles on arXiv and The World Health Organization (WHO). Chart-to-Text is another large figure captioning dataset that extends Obeid et al.'s data-to-text dataset. Its figures and captions predominantly consist of bar charts and are extracted from Pew Research \footnote{https://www.pewresearch.org/} and Statista. While more diverse than synthetic datasets, the type and amount of information provided in figure captions scraped from scientific papers and online sources can vary dramatically. Some captions only include limited, level 1 information which is insufficient for visually impaired readers; other captions include extraneous, level 4 information which no model or layperson can deduce without access to external knowledge or the content of the article. 
 
 \section{LineCap Dataset}

Unlike some previous works that focused on creating datasets of bar charts, we created a benchmark dataset of line charts, as they are the second-most commonly occurring type of charts in scientific publications (second to diagrams \cite{borkin2013}) and have more complex trends and patterns (i.e., level 3 information). We created a collection of line charts by taking a random sample of line plots from the SciCap dataset. SciCap \cite{scicap} used PDFFigures 2.0 \cite{pdffigures} to extract figures from scientific papers, then used a pre-trained classifier to identify the figure types -- with a reported accuracy of 86\%. As such, we manually inspected all sampled line plots to remove incorrectly cropped or classified figures. We also removed figures that were illegible due to poor quality, along with any multi-lined figures that were missing line labels or legends. Furthermore, to make the caption generation task easier for human annotators, we limited our scope to figures with at most five lines.

%% include figure of sample figures removed and kept from scicap?

\subsection{Caption Collection}
  Although Kim et al. \cite{kim:2021} provide guidelines for generating line chart captions, their guidelines are designed to enhance caption efficacy for sighted individuals. In this work, however, we focused on creating captions that enhance accessibility. As a result, following the recommendation of Lundgard et al. \cite{Lundgard:2022}, we created captions that describe overall trends and statistics (i.e., level 2 and 3 captions). Level 3 information is perceiver dependent and cannot be generated from data tables, without reference to the visualization \cite{Lundgard:2022}. As such, we used Amazon Mechanical Turk to crowd-source captions, unlike some previous works that used data tables to fill out template captions. According to Morash et al. \cite{morash:2015}, when collecting chart descriptions from novice web workers, using query templates is more suitable than providing a set of written guidelines. We used a modified version of Morash et al.'s template for line chart descriptions. More specifically, since we were focusing on high-level information, we did not keep any parts of the template that extracted low-level details (e.g., axis and line labels). We also added a question to capture additional information such as notable comparisons between the lines in a multi-lined figure. This format provides the necessary structure to ensure annotators provide all the high-level details that must be included in an accessible caption. However, it does not limit sentence structures and the types of trends that could be described.
 
%% include figure of our template in appendix (screenshot of MTurk setup)
To ensure no low-level content from the chart is referenced in the high-level captions, annotators were instructed to number the lines based on the order in the figure legend, and to refer to the lines by their numbers (i.e., Line 1 to 5) \footnote{Details can be found in our supplementary materials}. Annotators were also instructed to use the term {\fontfamily{qcr}\selectfont xlabel} to refer to the x-axis label and {\fontfamily{qcr}\selectfont ylabel} to refer to the y-axis label. Our objective in excluding any references to labels from the figure is two-fold. First, labels in scientific figures tend to include subscripts, equations, or special characters that cannot be found in a standard keyboard. By asking annotators not to use these labels in their descriptions, we avoid having to provide guidelines on transcribing these symbols. Second, this setup allows for training models that can be easily integrated with chart parsing models. Previous work on chart parsing can extract and classify text from figures \cite{poco:2017, chart-mining-survey}. The classified texts can then replace the standardized terminology in our high-level captions.

%% include figure of how lines are numbered (appendix)
 
Based on the results of an initial pilot study, we provided human annotators with a pictographic list of useful terminology for describing common line trends, along with video instructions that detail our Mechanical Turk interface. Because our task was writing intensive, we only accepted annotators from the following anglophone countries: Australia, Canada, New Zealand, UK, and USA. To ensure high quality, we only granted access to our annotation task, to those who first passed a qualification task. We also regularly inspected random samples of chart descriptions from each annotator.
 
The collected captions were processed to fix some spelling mistakes. We also followed a similar text normalization process to SciCap \cite{scicap}, whereby we replaced all references to axis values (e.g., 5.2, 10\%, 100k) with the token {\fontfamily{qcr}\selectfont \_value\_}. We provide access to both the annotations with axis values and the normalized annotations \footnote{\url{https://github.com/anita76/LineCapDataset}}.

\begin{figure}[h]
 \centering
 \includegraphics[width=\columnwidth]{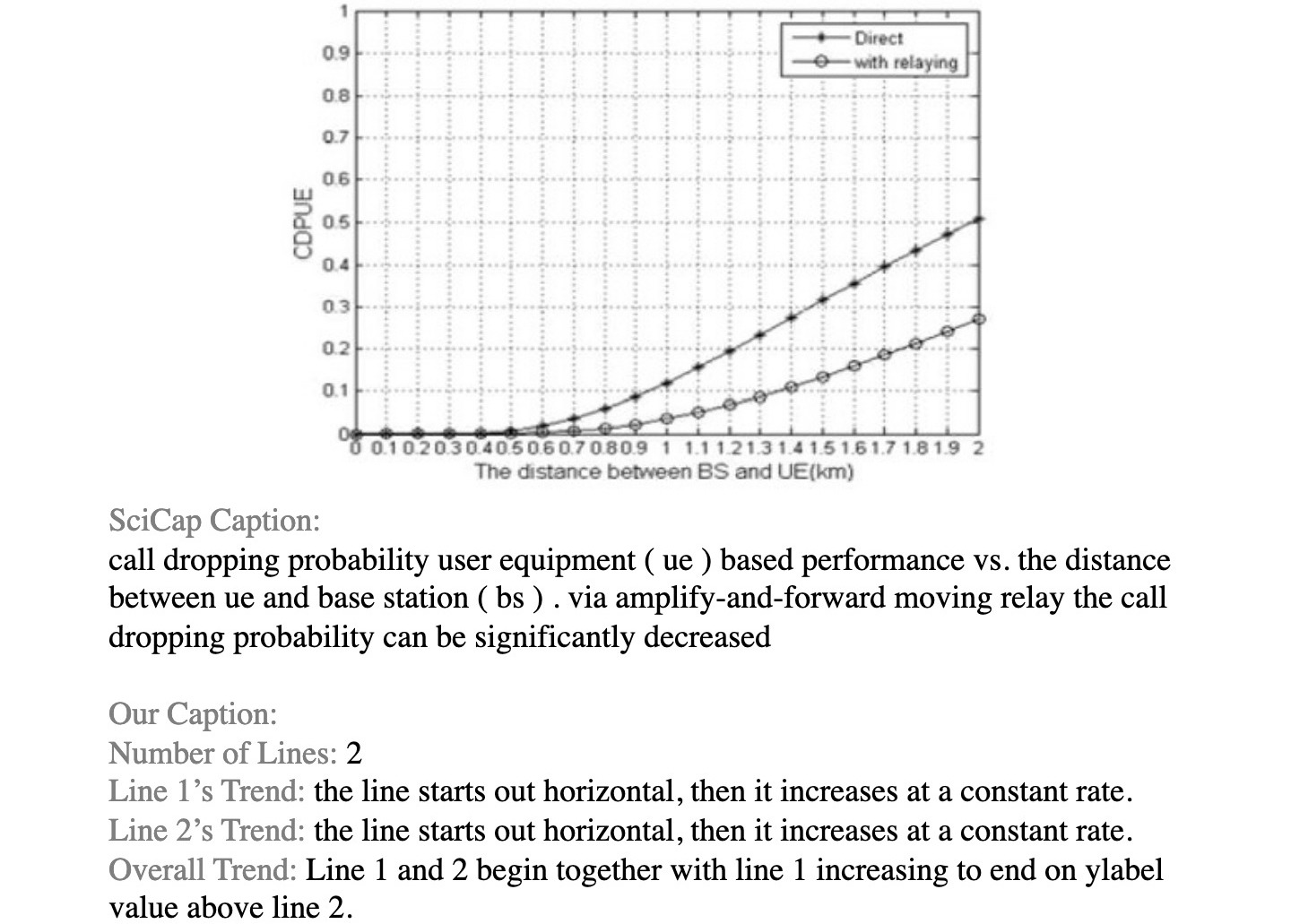}
 \caption{A sample figure-annotation pair from LineCap compared to the caption provided in the SciCap dataset.}
 \label{fig:sample-linecap-data}
 \vspace{-3mm}
\end{figure}
 
\begin{table}[h]
  \caption{Count and $\%$ of figures with the specified number of lines}
  \label{tab:figure frequency}
  \scriptsize%
	\centering%
  \begin{tabu}{%
	*{3}{c}%
	*{3}{c}%
	}
  \toprule
  number of lines  & count of figures & \% of total figures \\ 
  \midrule
  $1$              & $570$            & $16\%$              \\
  $2$              & $1025$           & $29\%$              \\
  $3$              & $829$            & $23\%$              \\
  $4$              & $796$            & $23\%$              \\
  $5$              & $308$            & $9\%$               \\
  \bottomrule
  \end{tabu}%
\end{table}
\vspace{-5mm}
 
\subsection{Dataset Analysis}
LineCap contains 3,528 figures, each with at least one human annotation that specifies: the number of lines in the figure; a separate description for the trend of each line in the figure; and, an overall chart description.
Most figures contain only a single annotation, but some figures have up to three different annotations. Our resulting dataset has a total of 3,964 annotations. The distribution of the number of lines in each figure can be found in \autoref{tab:figure frequency}. Figures most often have 2 lines, while the least common figures have 5 lines. An average description for a line trend is 14 words long, while an average description for the overall chart is 26 words long. Most annotations (both for individual lines and the overall chart), do not make any references to axis values from the chart. After pre-processing the descriptions (i.e., removing stop words and lemmatizing), the 10 most common words are:
line, increase, decrease, rate, ylabel, trend, xlabel, value, roughly, and constant. Excluding graph values, the pre-processed descriptions have a total of 925 unique words.

Individual line trends tend to describe the type of change (e.g., increase, decrease, constant) and the rate of change (e.g., increasing, decreasing or constant rate) of the line. Sometimes the trends also indicate the presence of noise, peaks, and troughs in the line. If a figure line is composed of multiple segments with different trends (e.g., first increasing then horizontal), descriptions tend to specify the trend of each segment in chronological order, but they do not indicate where exactly the change starts. The type of information provided to describe a figure's overall trend varies. When the figure has multiple lines, annotators tend to specify details such as: the relative rate of change of the lines, any notable line intersections, and the relative order of the lines along the y-axis. When the figure has a single line, the figure's overall trend either mirrors or is a summary of the description provided for the line's trend.

\begin{table}
    \caption{Average, mode, and maximum number of words and number of references to axis values for individual line and overall chart descriptions. The minimum number of words was 3 while the minimum number of axis value references was 0 for all description types.}
    \label{tab:word counts}
    \scriptsize%
    \centering%
    \begin{tabu}{%
      *{7}{c}%
	  *{7}{c}%
	  } 
        \toprule
         \multirow{2}{*}{\parbox{1.5cm}{\centering description type}} &\multicolumn{3}{c}{words per annotation} & \multicolumn{3}{c}{values per annotation}\\ 
        \cmidrule(lr){2-4}\cmidrule(lr){5-7}
                     & mean   & mode   & max   & mean  & mode & max \\ 
        \midrule
        line $1$      & $15.1$ & $6$    & $112$ & $0.6$ & $0$ &$16$ \\
        line $2$      & $14.4$ & $6$    & $123$ & $0.5$ & $0$ &$11$ \\
        line $3$      & $13.9$ & $6$    & $68$  & $0.4$ & $0$ &$11$ \\
        line $4$      & $13.1$ & $6$    & $96$  & $0.4$ & $0$ &$16$ \\
        line $5$      & $12.9$ & $6$    & $82$  & $0.4$ & $0$ &$6$  \\
        overall chart & $25.6$ & $16/17$& $123$ & $0.5$ & $0$ &$12$ \\
        \bottomrule
    \end{tabu}%
\end{table}

\section{Experiments}

To better understand LineCap's complexity, we set up a baseline deep learning model for predicting individual line trends. Training separate models for each of the figure lines is inefficient and does not scale well to figures with large number of lines. Furthermore, the knowledge required for generating descriptions for lines 1 through 5 is transferable between all these lines. As such, we built a model that generates descriptions for all the figure lines one at a time. We designed a two-staged deep learning pipeline that is comprised of two neural network models: a line count prediction and a caption generation model. First, the line count prediction model receives a figure as input and predicts its number of lines, $n$. Second, the pipeline iterates $n$ times. At every iteration, the caption generation model produces a description for the corresponding line, which is based on the following inputs: the figure image, the iteration number which indicates the line index, and a number indicating how many lines the figure has in total.

\begin{figure}
     \centering
     \includegraphics[width=\columnwidth]{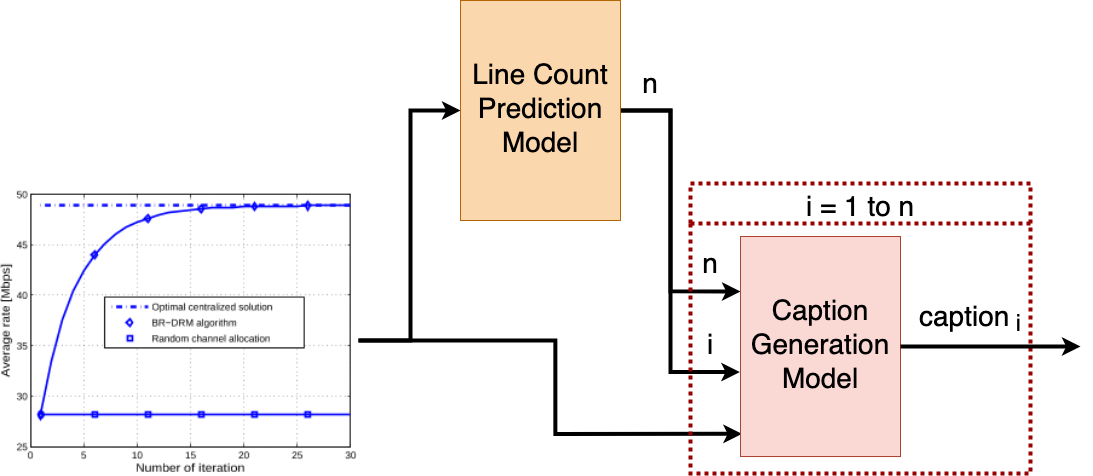}
     \caption{The deep learning pipeline for caption generation where $n$ is the predicted count of lines in the figure, $i$ is the line index which iterates from 1 to $n$, and $caption_{i}$ is the generated caption for line i.}
     \label{fig:overall-pipeline}
     \vspace{-5mm}
\end{figure}

The two models were implemented using a modified version of the PReFIL \cite{bimodalfusion} model. Despite its simple architecture compared to other figure question answering and captioning models, PReFIL achieves high accuracy performance on FigureQA \cite{figureqa} and DVQA \cite{dvqa}. Furthermore, unlike models such as FigJAM \cite{captionunits} and STL-CQA \cite{stlcqa}, it does not require any auxiliary annotations such as the figure texts and bounding boxes -- which our dataset does not supply.

\subsection{Line Count Prediction}

Our line count prediction model uses: a DenseNet \cite{densenet} to process the figure image; two fusion blocks \cite{bimodalfusion} for processing high and low level feature maps from the DenseNet; and, a neural network classifier that predicts the output. 

We used 80\% (2,822 figures) of our data for training, 10\% (353 figures) for validation and 10\% (353 figures) for testing. We trained the model using both our training data and the 40,000 line graphs in the training split of FigureQA. Adding FigureQA to our data, boosted the model's initial accuracy from about 70\% to about 90\%. Our model obtained a final accuracy of 94.05\% on our test set. This is lower than the 99.88\% accuracy on the line charts from FigureQA. To ensure that this discrepancy was not solely due to the much larger number of synthetic data from FigureQA, we also trained the model with our data and only a random sub-sample of 2,822 figures from FigureQA's training split. The accuracy values on the FigureQA dataset were still much greater than our dataset, thus illustrating the greater complexity of our real dataset compared to synthetic ones.

\begin{figure}
     \centering
     \includegraphics[width=\columnwidth]{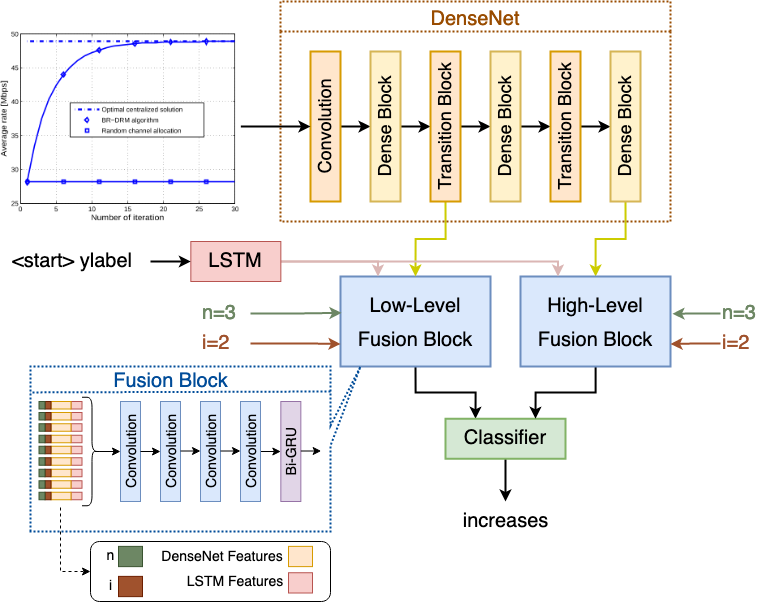}
     \caption{The architecture of the caption generation model. Several copies of the line index, $i$, and number of lines in the figure, $n$, are concatenated to the LSTM and DenseNet features, and are given as inputs to the fusion blocks. The line count prediction model has the same overall architecture but does not have an LSTM.}
     \label{fig:caption-model}
     \vspace{-1mm}
\end{figure}

\begin{table}
    \caption{Accuracy of the line count prediction model on the FigureQA and LineCap datasets when trained with the full FigureQA dataset and a sub-sample of the FigureQA dataset.}
    \label{tab:line-count-accuracy}
    \scriptsize%
    \centering%
    \begin{tabu}{%
      *{5}{c}%
	  *{5}{c}%
	  } 
        \toprule
        &\multicolumn{2}{c}{FigureQA} & \multicolumn{2}{c}{LineCap}\\ 
        \cmidrule(lr){2-3}\cmidrule(lr){4-5}
                     & Validation 1   & Validation 2   & Validation   & Test \\ 
        \midrule
        \parbox{1.5cm}{\centering Accuracy (\%) Full training}     & 99.88 & 99.88    & 94.90 & 94.05 \\
        \parbox{1.5cm}{\centering \vskip 0.2cm Accuracy (\%) Sub-sample training}     & 98.50 & 98.39    & 90.93 & 91.50 \\
        \bottomrule
    \end{tabu}%
    \vspace{-5mm}
\end{table}

\begin{figure*}[ht!]
    \centering
    \includegraphics[width=\textwidth]{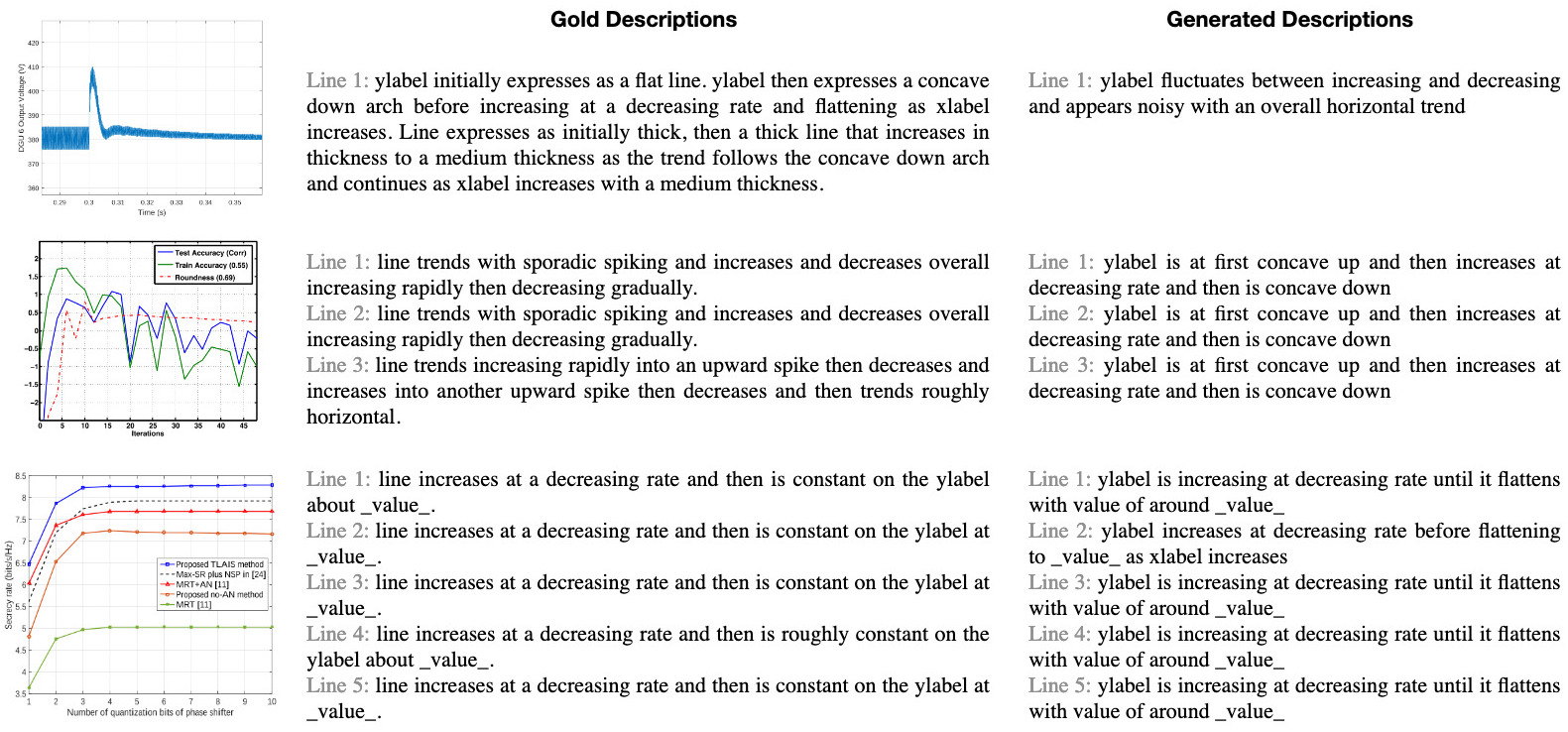}
    \caption{Sample gold and generated line descriptions for figures with one, three, and five lines.}
    \label{fig:generated-captions}
    \vspace{-3mm}
\end{figure*}

\subsection{Caption Generation}
Our caption generation model uses an LSTM to process the figure's caption one token at a time and predict the next token of the caption. We used beam search with $k=32$ to generate several captions and selected the caption with the highest weighted probability as the final prediction. We incorporated the scalar inputs (i.e., line index and number of lines in the figure) into the PReFIL model by passing several copies of them into the high and low level fusion blocks.

We report the model's performance in terms of three automated machine translation metrics: BLEU-4\footnote{We used NLTK's implementation of BLEU-4 with the smoothing 7 technique \cite{chen-cherry}} \cite{bleu}, CIDEr \cite{cider}, and BLEURT\cite{bleurt}. Additionally, to better assess the accuracy of generated captions, we randomly sampled 100 figures from our test split and asked Mechanical Turk workers to rank the accuracy of the descriptions for each of the figure lines on a scale of 1 to 5, where 1 is completely inaccurate and 5 is completely accurate.

\begin{table}[h]
  \caption{The caption generation model's performance on the test split of LineCap. Accuracy is reported for a sub-sample of 100 figures. All scores are reported when the ground-truth line count number is used in the caption generation model. Scores that use the predicted line count number are in the supplementary materials.}
  \label{tab:caption results}
  \scriptsize%
	\centering%
  \begin{tabu}{%
	*{5}{c}%
	*{5}{c}%
	}
  \toprule
  figure with  & BLEU-4  & CIDEr  & BLEURT-20 & Accuracy \\ 
  \midrule
  1 line       & $0.366$  & $1.173$  & $0.511$ &  \textbf{3.43}   \\
  2 lines      & $0.418$  & $1.096$  & $0.524$ &  3.13   \\
  3 lines      & $0.433$  & $1.244$  & $0.521$ &  3.03   \\
  4 lines      & $0.443$  & \textbf{1.458}  & 0.516 & 3.22    \\
  5 lines      & \textbf{0.455}  & $1.018$   & \textbf{0.529} &  3.36   \\
  \midrule
  all figures     & $0.433$  & $1.229$  & $0.522$  & 3.20  \\
  \bottomrule
  \end{tabu}%
  \vspace{-3mm}
\end{table}

Based on human evaluations, descriptions of single-lined figures are the most accurate. The model generally does well on simple trends but struggles with more complex trends and multi-lined charts. For multi-lined charts, we observe that the model mostly repeats the same description for all of the figure lines even if the lines do not display the same trend (e.g., second example in \autoref{fig:generated-captions}). Some of the higher accuracy scores for multi-lined charts are due to all the figure lines having the same overall trend, as is the case with the last example in \autoref{fig:generated-captions}. This could be because the model is unable to learn the correlation between the line number indices and the lines in the figure. Additional figure annotations such as bounding boxes and labels could help guide the model in learning this information.

We also calculate correlation coefficients between the automated metrics and human evaluation. BLEURT has the highest correlation of 0.45. CIDEr and BLEU-4 have correlation coefficients of only 0.26 and 0.25, respectively, suggesting that these metrics are likely not suitable for effectively comparing figure captioning models.

\section{Conclusion}
We created LineCap, a novel dataset of line charts for figure captioning models. We subsequently established baseline line count and caption prediction performances. Through this work, we gathered the following insights and areas of future research toward automated captioning of data visualizations using deep learning models:

\begin{itemize}[leftmargin=5.5mm]
    \setlength\itemsep{1mm}
    \item Future work should aim to create datasets similar to LineCap for other chart types. Creators of datasets for automated figure captioning should focus on gathering real figures and human-written captions. Novice web-workers should be provided with further guidance on how to describe figures in order to ensure accuracy. Future research could also benefit from investigating how visually impaired audiences perceive the descriptions written by sighted individuals.

    \item Previous works \cite{kilickaya-2017} have investigated the limitations of automated metrics for evaluating \textit{image} captioning models, and have proposed guidelines for proper assessment of these models. \textit{Figure} captions have additional nuances that are not common in image captions. For instance, while the order in which objects in an image are described is not important, changing the order of trends in a line would lead to inaccurate descriptions. Such nuances warrant similar investigations toward identifying suitable metrics for evaluating figure captioning models. Our experiments have shown that some common automated evaluation metrics do not correlate well with human evaluation. To develop state-of-the-art figure captioning models, future research should first identify suitable automated evaluation metrics for this task.

    \item Deep learning models for figure captioning can benefit from incorporating additional intermediate prediction tasks from previous research on chart processing and analysis \cite{chart-mining-survey}. Tasks such as segmenting the lines or extracting data values from the figures could be particularly useful for distinguishing between the different lines in a multi-lined figure. Training intermediate tasks on large, synthetic datasets and fine-tuning on smaller, real datasets could significantly boost performance without the costs of creating large, crowd-sourced datasets.
\end{itemize}

\acknowledgments{
We would like to thank Zhutian Chen and the anonymous reviewers for their valuable feedback. We would also like to thank Mechanical Turk workers for their contributions to the LineCap dataset. This work was supported by a grant from the Harvard Data Science Initiative. Some of the computations in this work were run on the FASRC Cannon cluster at Harvard University.}

\bibliographystyle{abbrv-doi}

\bibliography{template}
\end{document}